\documentclass[runningheads]{llncs}
\usepackage[T1]{fontenc}
\usepackage{multirow,tabularx}
\usepackage{mathrsfs}
\usepackage{hyperref}
\usepackage{graphicx}
\usepackage{amssymb}
\usepackage{amsmath}
\usepackage{floatrow,makecell}
\usepackage{url}
\usepackage{comment}
\usepackage{tablefootnote}
\usepackage{xcolor}

\begin{document}

\title{Towards Deployable OCR Models for Indic Languages}
\titlerunning{Towards Deployable OCR Models for Indic Languages}

\author{Minesh Mathew\orcidID{0000-0002-0809-2590} \and Ajoy Mondal\orcidID{0000-0002-4808-8860} \and C V Jawahar\orcidID{0000-0001-6767-7057}}

\authorrunning{Mathew \emph{et al.}}

\institute{CVIT, International Institute of Information Technology, Hyderabad, India\\
\email{minesh.mathew@gmail.com}\\
\email{\{ajoy.mondal,jawahar\}@iiit.ac.in}}

\maketitle             
\begin{abstract}

The difficulty of reliably extracting characters had delayed the character recognition solutions (or OCRs) in Indian languages. Contemporary research in Indian language text recognition has shifted towards recognizing text in word or line images without requiring sub-word segmentation, leveraging Connectionist Temporal Classification (CTC) for modeling unsegmented sequences. The next challenge is the lack of public data for all these languages. And there is an immediate need to lower the entry barrier for startups or solution providers. With this in mind, (i) we introduce \textit{Mozhi} dataset, a novel public dataset comprising over 1.2 million annotated word images (equivalent to approximately 120 thousand text line images) across 13 languages. (ii) 
We conduct a comprehensive empirical analysis of various neural network models employing CTC across 13 Indian languages. (iii) We also provide APIs for our OCR models and web-based applications that integrate these APIs to digitize Indic printed documents. We compare our model's performance with popular publicly available OCR tools for end-to-end document image recognition. Our model outperform these OCR engines on 8 out of 13 languages. The code, trained models, and dataset are available at~\url{https://cvit.iiit.ac.in/usodi/tdocrmil.php}.

\end{abstract}

\section{Introduction} \label{sec:intro}

\begin{figure}
\centerline{
\includegraphics[height=0.4\textwidth,width=1.0\textwidth]{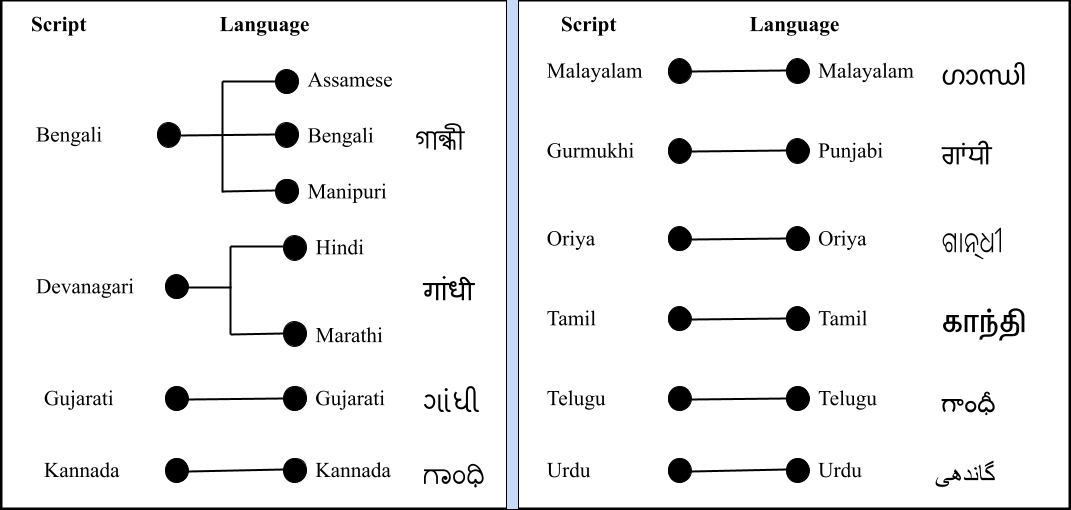}}
\caption{We explore printed text recognition across 13 Indian languages, covering ten unique scripts. Although many languages share a common alphabet, their scripts vary, with exceptions like Hindi and Marathi. The last column shows the name "Gandhi" in all ten scripts.} \label{fig:gandhi_words}
\end{figure}

\begin{figure}[!h]
\centerline{
\includegraphics[height=0.18\textwidth,width=0.23\textwidth]{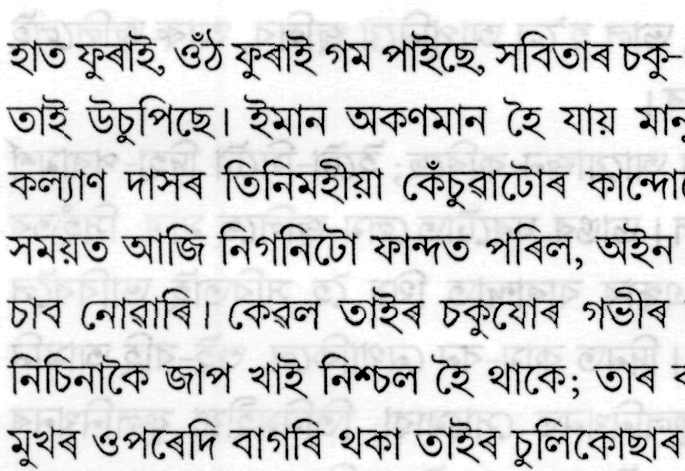}
\hspace{0.0001\textwidth}
\includegraphics[height=0.18\textwidth,width=0.23\textwidth]{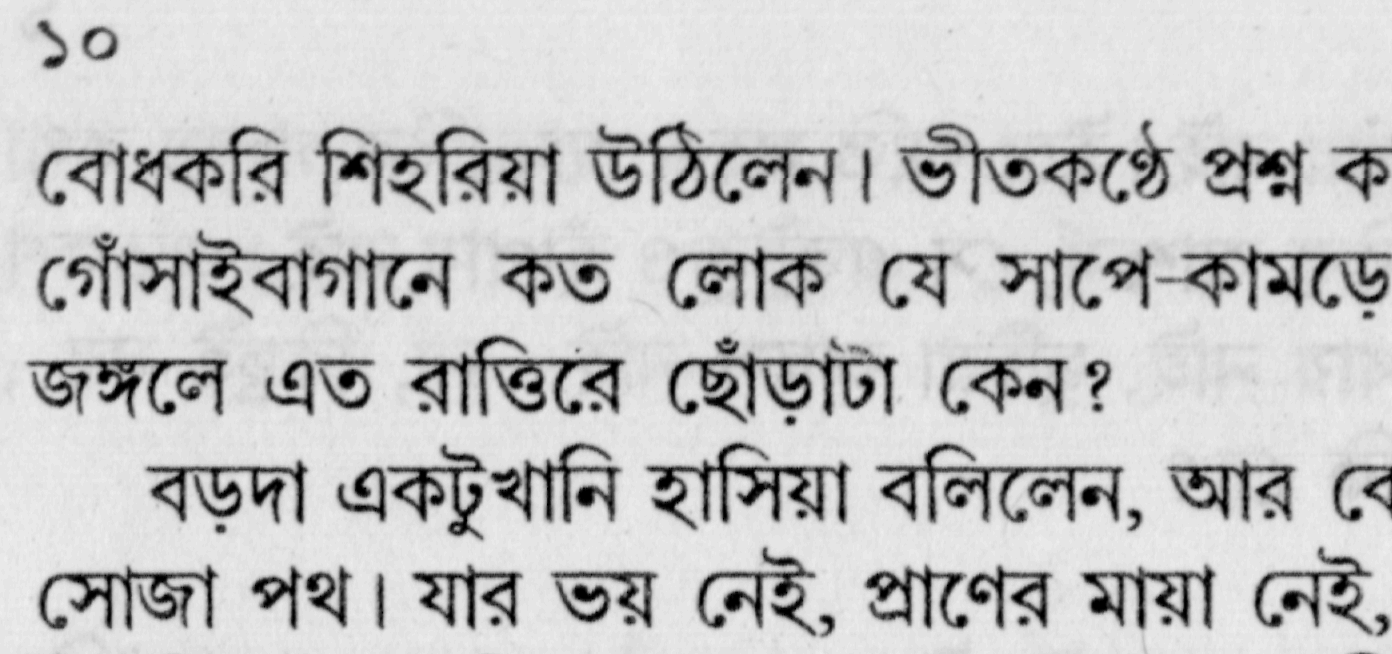}
\hspace{0.0001\textwidth}
\includegraphics[height=0.18\textwidth,width=0.23\textwidth]{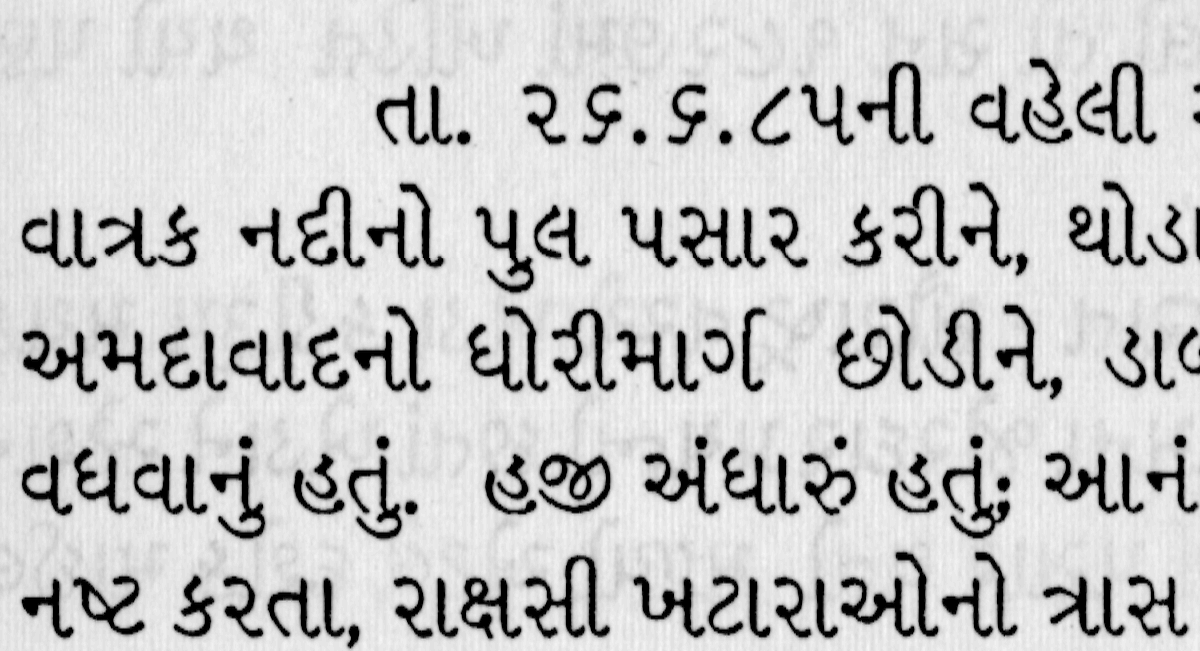}
\hspace{0.0001\textwidth}
\includegraphics[height=0.18\textwidth,width=0.23\textwidth]{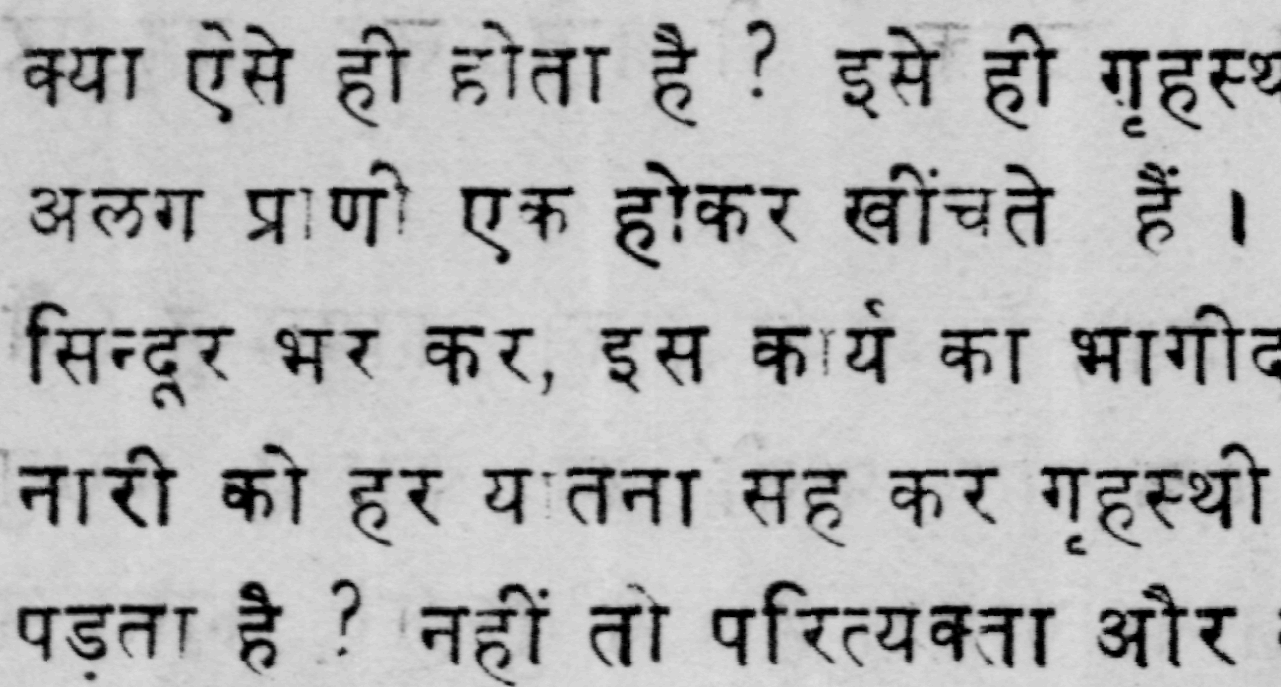}}
\vspace{0.01\textwidth}
\centerline{
\includegraphics[height=0.18\textwidth,width=0.23\textwidth]{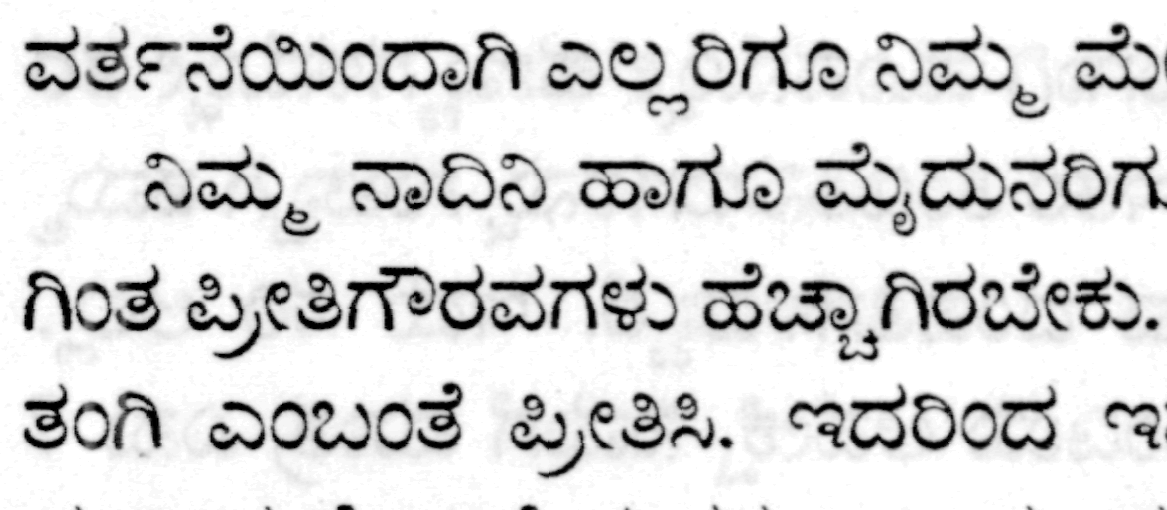}
\hspace{0.0001\textwidth}
\includegraphics[height=0.18\textwidth,width=0.23\textwidth]{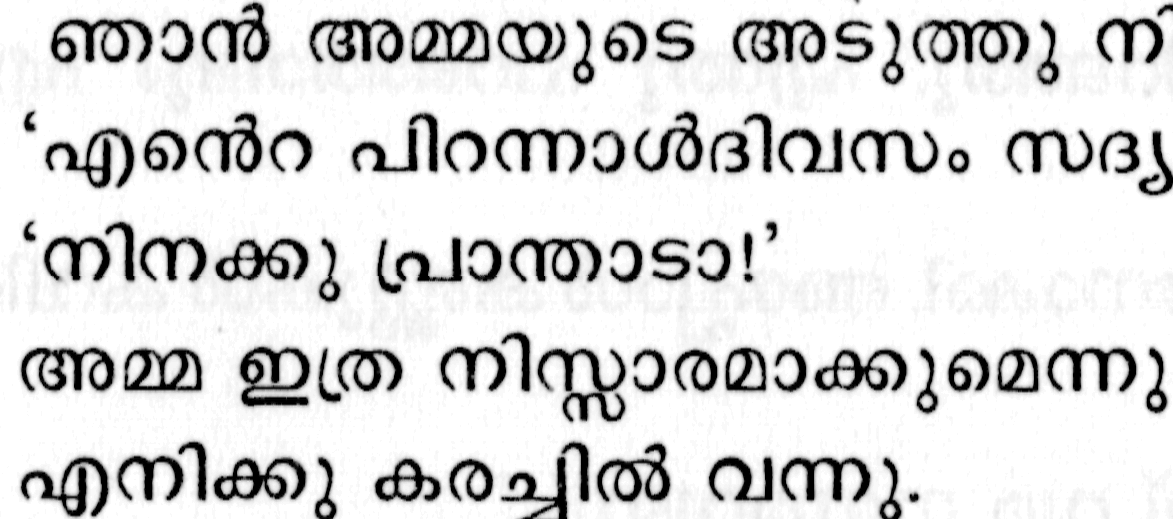}
\hspace{0.0001\textwidth}
\includegraphics[height=0.18\textwidth,width=0.23\textwidth]{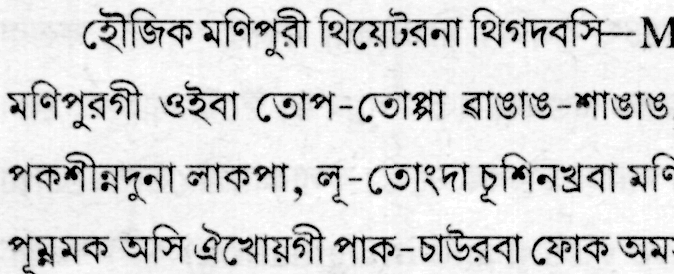}
\hspace{0.0001\textwidth}
\includegraphics[height=0.18\textwidth,width=0.23\textwidth]{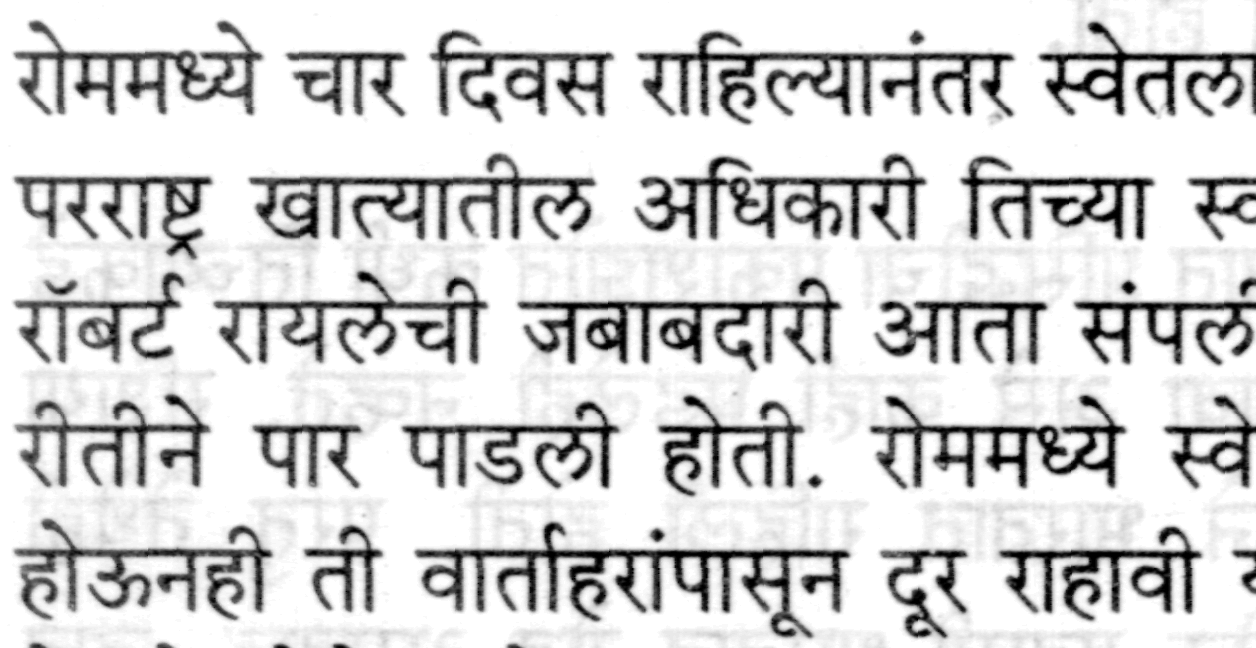}}
\vspace{0.01\textwidth}
\centerline{
\includegraphics[height=0.18\textwidth,width=0.23\textwidth]{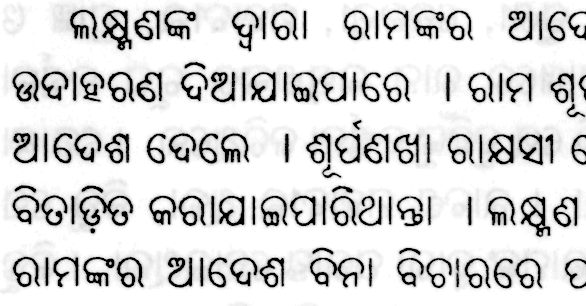}
\hspace{0.0001\textwidth}
\includegraphics[height=0.18\textwidth,width=0.23\textwidth]{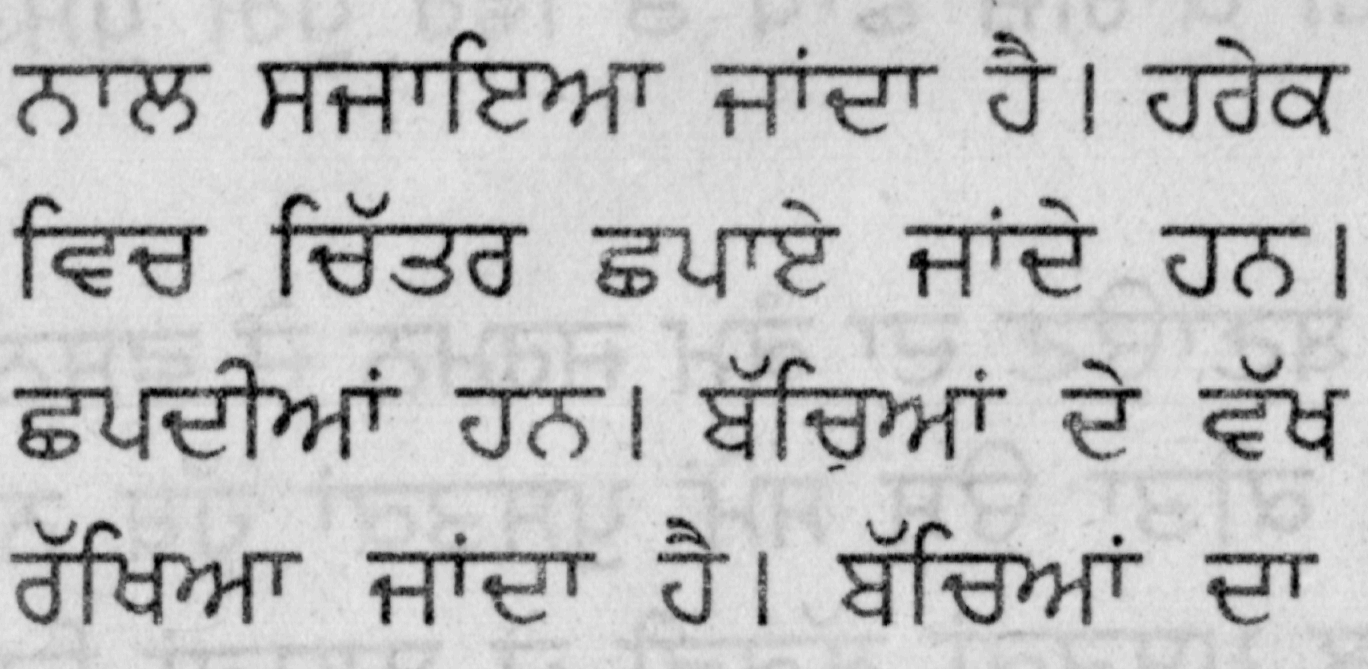}
\hspace{0.0001\textwidth}
\includegraphics[height=0.18\textwidth,width=0.23\textwidth]{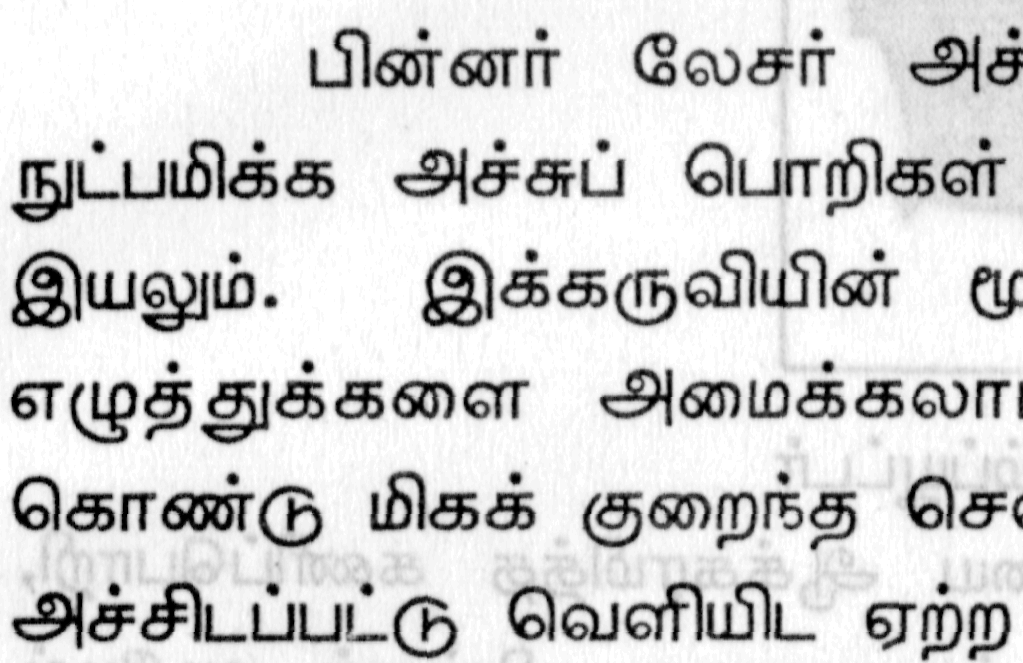}
\hspace{0.0001\textwidth}
\includegraphics[height=0.18\textwidth,width=0.23\textwidth]{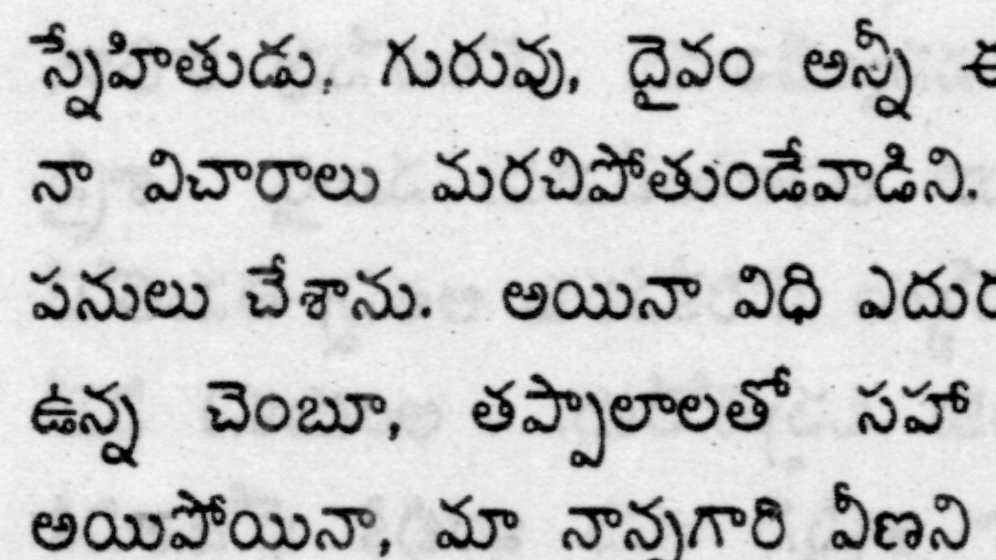}}
\vspace{0.01\textwidth}
\centerline{
\includegraphics[height=0.16\textwidth,width=0.8\textwidth]{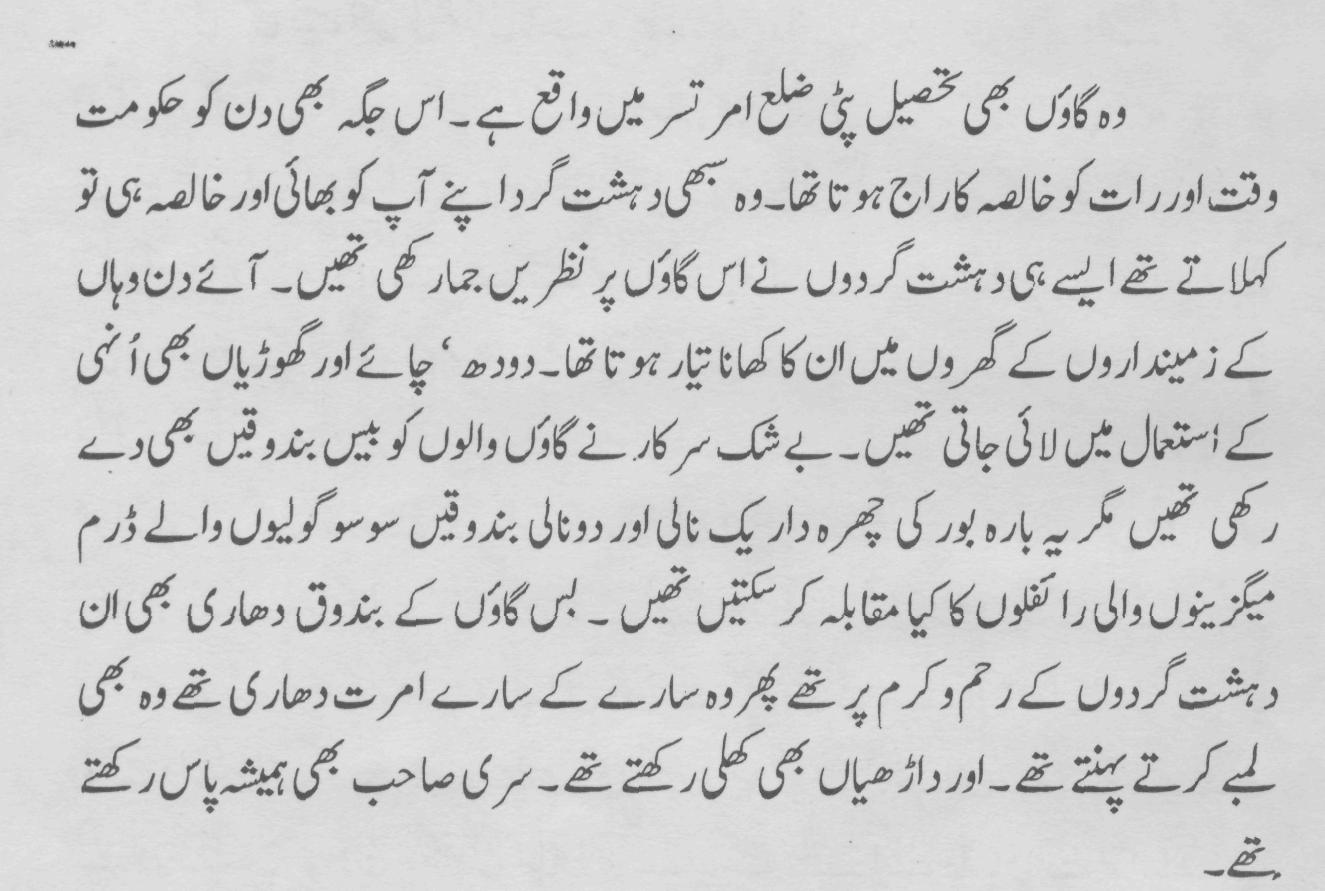}}
\caption{Shows a few sample of cropped images of each of 13 languages from our \textit{Mozhi} dataset.}
\label{fig:sample_crop}
\end{figure}

Text recognition faces challenges related to language/script, text rendering, and imaging methods. This study concentrates on recognizing printed text in Indian languages, particularly on text recognition alone, assuming cropped word or line images are provided. The 2011 official census of India~\cite{census2011} lists 30 Indian languages with over a million native speakers, 22 of which are recognized as official languages. These languages belong to three language families: \textit{Indo-European}, \textit{Dravidian}, and \textit{Sino-Tibetan}. Our focus is on text recognition in 13 official languages: \textit{Assamese, Bengali, Gujarati, Hindi, Kannada, Malayalam, Manipuri, Marathi, Oriya, Punjabi, Tamil, Telugu, and Urdu}. While some share linguistic similarities, their scripts are distinct, with Devanagari script used in Hindi and Marathi and Bengali script in Bengali, Assamese, and Manipuri, among others. Our study explores printed text recognition across 13 Indian languages, representing ten scripts. Fig.~\ref{fig:gandhi_words} illustrates "Gandhi" written in these ten scripts. At the same time, Fig.~\ref{fig:sample_crop} depicts a sample of cropped images from 13 languages from our newly created \textit{Mozhi} dataset. The APIs corresponding to our developed models are integrated into Bhashini\footnote{\url{https://bhashini.gov.in/}} for public use. However, we are continuously working on including the remaining low-resource languages --- \textit{Bodo, Dogri, Kashmiri, Konkani, Maithili, Nepali, Sanskrit, Santali, and Sindhi} --- to cover all twenty-two languages of India.

Efforts to develop OCRs for Indian scripts began in the 1970s but faced challenges in scaling across languages and achieving satisfactory results across diverse document types until recently~\cite{mahabala,bbc_bengali_1998,MOCR2011}. Challenges such as script intricacies, linguistic diversity, and limited annotated data hindered progress in Indian language OCR. The adoption of Connectionist Temporal Classification (CTC), initially successful in speech transcription, revolutionized text recognition across various forms, including handwritten~\cite{ctc_handwritten}, printed~\cite{breuelICDAR,BLSTM_ICPR}, and scene text~\cite{su_rnn_ctc,crnn_arxiv}. Popular open-source OCR tools like \textit{Tesseract}~\cite{tesseract_github}, \textit{EasyOCR}~\cite{easyocr}, and \textit{ocropy}~\cite{ocropus_github} now leverage CTC-based models, enabling recognition of word or line images without sub-word segmentation.

Segmenting words into sub-word units presents a significant challenge for Indian languages compared to English~\cite{BLSTM_ICDAR}. Developing Indian language recognizers is further complicated by the intricate relationships between script glyphs, language text, and machine representation. In the script, the atomic unit is an isolated symbol (glyph), while in the language, it's an \textit{Akshara} or an orthographic syllable. Machine text representation uses \textit{Unicode} points. An \textit{Akshara} can comprise multiple glyphs, and a sequence of multiple \textit{Unicode} points can represent an \textit{Akshara}. Splitting text at \textit{Akshara}s and mapping them to \textit{Unicode} sequences necessitates language and script knowledge~\cite{BLSTM_ICDAR,minesh_das}. Therefore, adopting CTC-based sequence modeling has become the standard approach for Indian language OCR~\cite{BLSTM_ICDAR,adnanICDAR_urdu_2013,PraveenDAS}. This approach directly maps features from word or line images to target \textit{Unicode} sequences, eliminating the need for explicit alignment during training. Our study offers a comprehensive empirical analysis of various design considerations in developing a CTC-based printed text recognition model for Indian languages.

\begin{figure}[!h]
\centerline{
\includegraphics[height=0.55\textwidth,width=1.0\textwidth]{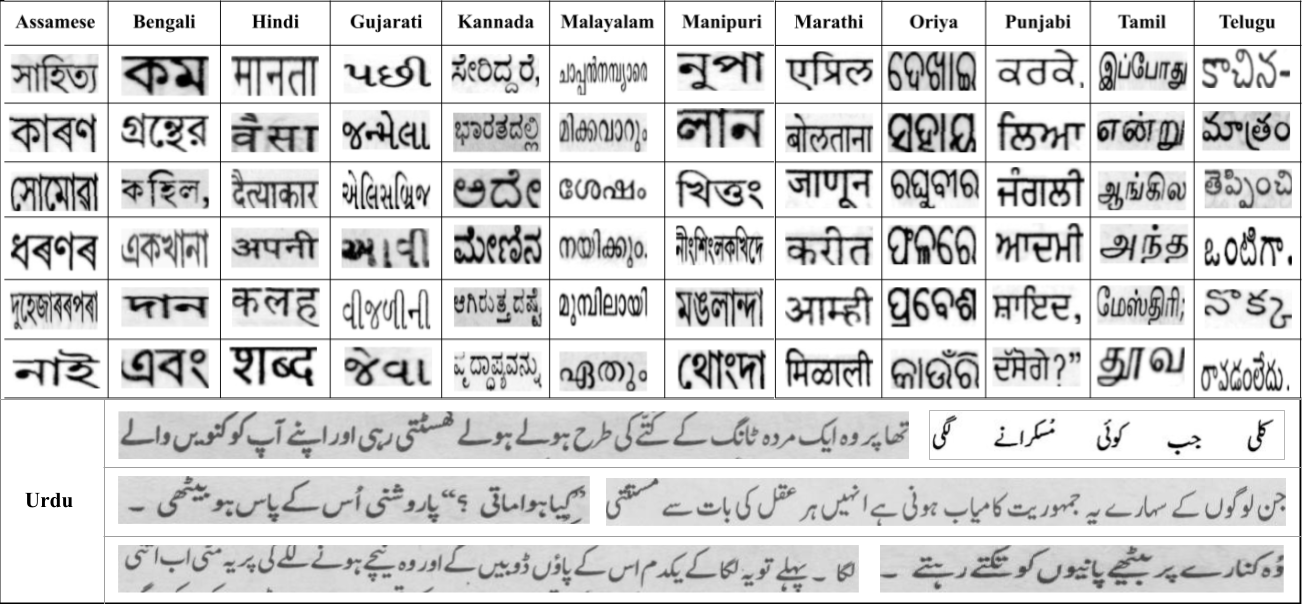}}
\caption{A few sample of word level images from our \textit{Mozhi} dataset.}
\label{fig:sample_word}
\end{figure}

Our contributions are the following:

\begin{itemize}
\item We introduce a new public dataset \textit{Mozhi} for text recognition in 13 Indian languages, comprising cropped line and word segments with corresponding ground truth for all languages except Urdu. With over 1.2 million annotated word images, this dataset is the largest for text recognition in Indian languages (refer Table~\ref{table:public_dataset_Stats} and Fig.~\ref{fig:sample_word}).

\item We empirically compare the performance of four types of CTC-based text recognition methods across 13 official languages of India, varying in feature extraction and sequence encoding. Additionally, we assess word level and line level recognition models. 

\item We develop end-to-end page level OCR systems by integrating our best text recognition models with existing line and word segmentation tools. These systems outperform \textit{Tesseract5}~\cite{tesseract_github} and \textit{Google Cloud Vision OCR}~\cite{google_cloud_vision} for 8 out of 13 languages (refer Table~\ref{table:page_level_end_to_end}).

\item Offer APIs for our OCR models and web-based applications that seamlessly integrate these APIs to digitize Indic printed documents.  

\end{itemize}

\section{Related Work} 

Current OCRs for Indian scripts mainly rely on segmentation-free approaches, which directly produce a label sequence from word or line images. Sankaran \emph{et al.}~\cite{BLSTM_ICPR} introduced CTC-based sequence modeling for printed text recognition in Indian languages. Their method utilizes an RNN encoder and CTC transcription to map features extracted from Devanagari word images to class labels. Profile-based features~\cite{manmathaicdar} extracted using a 25 $\times$ 1 sliding window are employed. Initially, the model maps \textit{Akshara}s to class labels and uses rule-based mapping to Unicode. In a subsequent work~\cite{BLSTM_ICDAR}, they directly map feature sequences from word images to \textit{Unicode} sequences, eliminating the need for rule-based \textit{Akshara} to \textit{Unicode} mapping.

The introduction of the CTC-based transcription method marked a significant advancement in Indic scripts, particularly by overcoming the challenge of sub-word segmentation. Directly transcribing word images into machine-readable \textit{Unicode} sequences also eliminated the need for language-specific rules to map latent output classes to valid \textit{Unicode} sequences. Krishnan \emph{et al.}~\cite{PraveenDAS} utilized profile-based features and a CTC-based model similar to~\cite{BLSTM_ICDAR} for recognizing seven Indian languages. Their evaluation on a large test set per language demonstrated the effectiveness of a unified framework employing CTC transcription for multilingual text recognition, eliminating the necessity for language or script-specific modules.

Hasan \emph{et al.}~\cite{adnanICDAR_urdu_2013} proposed an RNN+CTC model for printed Urdu text recognition, directly generating Unicode sequences from text line images. Utilizing a 30 $\times$ 1 sliding window for raw pixel feature extraction, their method yielded promising outcomes. Similarly, our prior work\cite{minesh_das} centered on multilingual OCR for 12 Indian languages and English, employing a two-stage system with a script identification module and a recognition module. Chavan \emph{et al.}~\cite{blstm_mdlstm_2017_7languages} compared RNN and multidimensional RNN (MDRNN) encoders with CTC transcription. They found the MDRNN encoder outperformed the RNN encoder, using HOG features with the former and raw pixels with the latter. Another study achieved over 99\% character/symbol accuracy for Bengali script recognition~\cite{bengali_blstm_2019_bbc} using an RNN+CTC model. Kundaikar and Pawar~\cite{Devanagari_blstm_2020} explored the robustness of CTC-based Devanagari OCR to font and size variations. At the same time, Dwivedi \emph{et al.}~\cite{sanskrit_ocr_saluja} achieved a character/symbol error rate under 3\% for Sanskrit recognition using an encoder-decoder model. These findings, particularly the reliance on CTC transcription, motivate our comprehensive empirical study comparing various encoder types and features for both line and word recognition in Indian languages.

\section{Mozhi Dataset}\label{sec:public dataset}

To our knowledge, no extensive public datasets are available for printed text recognition in Indian languages. Early studies often utilized datasets with cropped characters or isolated symbols for character classification~\cite{kannada_kunte_2007,jawaharBilingual}. Later research relied on either internal datasets or large-scale synthetically generated samples for word or line level annotations~\cite{BLSTM_ICPR,PraveenDAS,minesh_das,blstm_mdlstm_2017_7languages,mohit_urdu,sanskrit_ocr_saluja,Devanagari_blstm_2020,adnanICDAR_urdu_2013}. While recent efforts have introduced public datasets for Hindi and Urdu, they typically contain a limited number of samples intended solely for model evaluation~\cite{minesh_das,mohit_urdu}. However, due to variations in training data among these studies, comparing methods can be challenging. To address the scarcity of annotated data for training printed text recognition models in Indian languages, we introduce the \textit{Mozhi} dataset. This public dataset encompasses both line and word level annotations for all 13 languages examined in this study. It includes cropped line images, corresponding ground truth text annotations for all languages, and word images and ground truths for all languages except Urdu. With 1.2 million word annotations (approximately 100,000 words per language), it is the largest public dataset of real word images for text recognition in Indian languages. For each language, the line level data is divided randomly into training, validation, and test splits in an 80:10:10 ratio, with words cropped from line images forming corresponding splits for training, validation, and testing. Table~\ref{table:public_dataset_Stats} shows statistics of \textit{Mozhi}.

\begin{table}[!h]
\begin{center}
\begin{tabular}{|l|l||r|r||r|r||r|r|} \hline 
\textbf{Script} &\textbf{Language} &\multicolumn{2}{c||}{\textbf{Train}} &\multicolumn{2}{c||}{\textbf{Validation}}                &\multicolumn{2}{c|}{\textbf{Test}} \\\cline{3-8}
    & &\textbf{Lines} &\textbf{Words} &\textbf{Lines} &\textbf{Words} &\textbf{Lines} &\textbf{Words} \\\hline\hline
Bengali    &Assamese &9566 &79959 &1196  &9945  &1196  &10146 \\
Bengali    &Bengali  &7579 &80113 &948    &9787  &947   &10113 \\
Gujarati   &Gujarati &8632 &79910 &1080 &10016  &1079  &10090 \\
Devanagari &Hindi    &6525 &79762 &816  &10114  &816   &10173 \\
Kannada    &Kannada  &13462 &80085 &1683  &10088  &1683  &9838 \\
Malayalam &Malayalam &15112 &80146 &1889  &9893  &1889  &9980  \\
Bengali &Manipuri &9765  &79691 &1221  &10254 &1221  &10061 \\
Devanagari &Marathi  &8380  &80151 &1048  &10005 &1048  &9855  \\
Oriya &Oriya  &8260  &79945 &1033  &10089 &1033  &9994  \\
Gurumukhi &Punjabi  &6726  &79931 &841   &10036 &841  &10038 \\
Tamil &Tamil &16074 &80022 &2010  &10021 &2009  &9974  \\
Telugu &Telugu   &12722 &80337 &1591  &9811  &1590  &9876  \\
Nastaliq  &Urdu  &9100  &-    &1138  &-  &1137  &-   \\\hline
\end{tabular}
\end{center}
\caption{Statistics for the new \textit{Mozhi} dataset, a public resource for recognizing printed text in cropped words and lines, reveal over 1.2 million annotated words in total. Notably, only cropped lines are annotated for Urdu.}\label{table:public_dataset_Stats}
\end{table}

\begin{figure}[!ht]
\centerline{
\includegraphics[height=0.6\textwidth,width=1.0\textwidth]{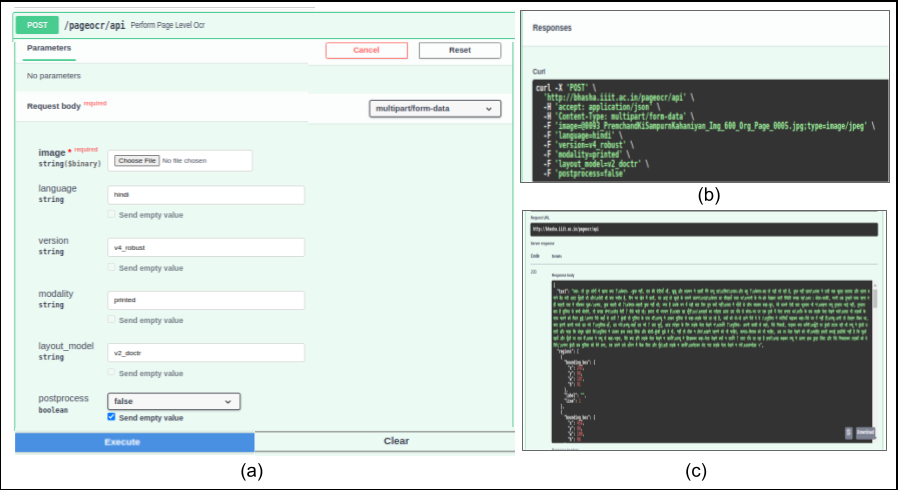}}
\caption{Shows screen shot of our web-based APIs to digitize Indic printed documents.}
\label{fig:api}
\end{figure}

\section{APIs and Web-based Applications}

We develop APIs for page level recognition models across 13 languages and built a web-based application available at~\url{https://ilocr.iiit.ac.in/fastocr/} that integrates these APIs for digitizing printed documents in Indic languages. Fig.~\ref{fig:api} illustrates the steps for utilizing our web-based APIs to digitize Indic printed documents. Users can upload a document image, select the language, OCR model version, layout version, and execute to obtain OCR output.           
\section{Text Recognition using CTC Transcription} \label{sec: method}

\begin{figure}[!ht]
\centerline{
\includegraphics[height=0.35\textwidth,width=1.0\textwidth]{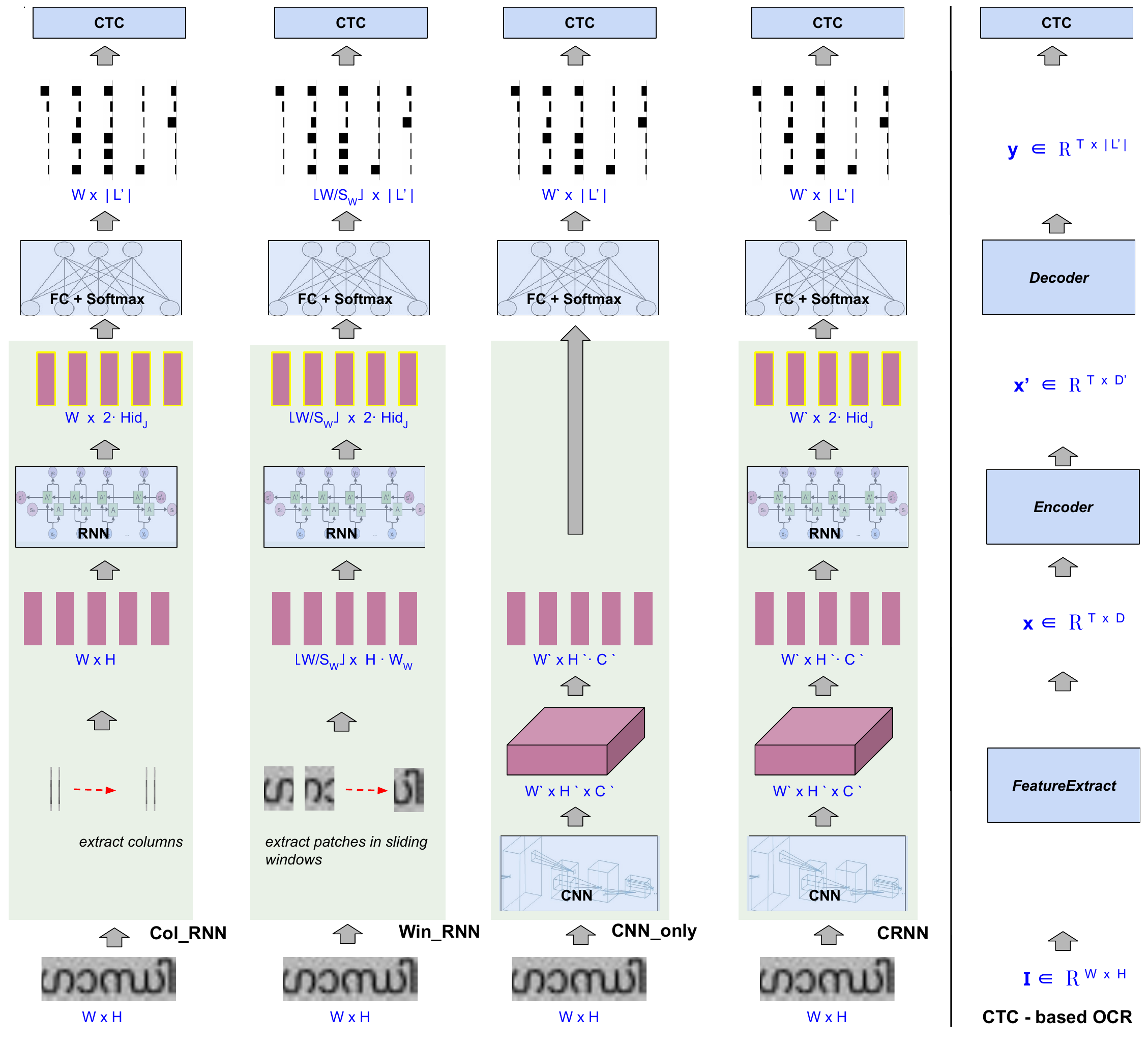}}
\caption{We examine four CTC-based text recognition methods --- Col\_RNN, Win\_RNN, CNN\_only, and CRNN, distinguished by their feature extraction and sequence encoding. $W$ and $H$ represent the width and height of the input image $I$, respectively. $|L^\prime|$ indicates the number of class labels, including the \textit{blank} label. $Hid_j$ signifies the number of hidden units in the last RNN layer. In the case of Win\_RNN, $W_W$, and $S_W$ denote the width and step size of the sliding window, respectively.}\label{fig:different_configurations}
\end{figure}

Given an input image $I$ containing a word or a line, text recognition involves converting the text on the image into a machine-readable format. We frame this task as a sequence modeling problem utilizing CTC. The input comprises a sequence of features $\mathbf{x}=x_1,x_2,...,x_T$, where $x_t \in \mathbb{R}^D$ is extracted from the image $I$. The output is a sequence of class labels $\mathbf{l}=l_1,l_2,...,l_N$, where $l_n \in {L}$ and $L$ represents the output alphabet, i.e., the set of unique class labels. In our scenario, ${L}$ corresponds to all \textit{Unicode} code points we aim to recognize. We adopt an encoder-decoder interpretation of the CTC framework, as described in~\cite{distil_ctc}.

\subsection{Extracting Feature Sequence} \label{sec:method_feat_extraction}

Graves \emph{et al.}~\cite{ctc_icml_2006} introduced CTC for speech-to-text transcription, employing a sliding window method to extract features from the time axis of the speech signal. They used a window size of 10 milliseconds (ms) and a step size of 5 ms, extracting a fixed-size feature vector termed a time-step or a frame at each instance of the sliding window. However, grey-scale images represent 2D scalar-valued spatial signals in contrast to speech signals. Thus, approaches employing CTC for text transcription from images typically extract features along the horizontal axis of the image~\cite{BLSTM_ICDAR,adnanICDAR_urdu_2013,crnn_arxiv}. We follow a methodology similar to that outlined in~\cite{BLSTM_ICDAR,adnanICDAR_urdu_2013,crnn_arxiv}, where feature vectors in the input sequence $\mathbf{x}$ represent horizontal segments of the image. Each instance of the input sequence is referred to as a time-step or a frame, consistent with the original approach~\cite{ctc_icml_2006}. The horizontal span of a frame varies depending on the feature extraction method. The feature sequence, $\mathbf{x}$, is extracted in alignment with the script direction. Specifically, for languages other than Urdu, features are extracted from left to right, whereas they are extracted in the opposite direction for Urdu. In summary, given a document image $I \in \mathbb{R}^{W\times H}$ (grey-scale), the feature sequence is obtained as follows:
\begin{equation}
     \mathbf{x}  \in  \mathbb{R}^{T\times D}  = FeatureExtract (I).  
\end{equation}
   
\paragraph{\textbf{Encoder:}}\label{sec:method_encoder}

The sequence encoder's task is to transform the input sequence $\mathbf{x}$ into an encoded representation $\mathbf{x^{\prime}} \in \mathbb{R}^{T\times D^{\prime}}$, where $D^{\prime}$ represents the encoding size --- i.e., the fixed dimensional to which each feature vector is encoded. 
\begin{equation}
\mathbf{x^{\prime}} \in \mathbb{R}^{T\times D^{\prime}} = Encoder (\mathbf{x}).
\end{equation}
In this work, we explore several encoder configurations --- Col\_RNN, Win\_RNN, CNN\_only, and CRNN for feature extraction.  

\paragraph{\textbf{Decoder:}} \label{sec:method_decoder}

The encoded features $\mathbf{x^\prime}$ undergo a linear projection layer followed by Softmax normalization, aligning their size with the number of output classes. This procedure, resembling the decoding phase of CTC as interpreted in~\cite{distil_ctc}, extends the original output alphabet $L$ with an extra label for blank, denoted as $\sim$. The blank label signifies instances where no label is assigned to an input. Softmax normalization at each time step yields class conditional probabilities, forming the posterior distribution over the classes. Essentially, given the sequence of encoded features,
\begin{equation}
    \mathbf{y} \in \mathbb{R}^{T\times L^{\prime} } = Decoder (\mathbf{x^\prime }),
\end{equation}
where each $y_t \in R^{L^{\prime}}$ represent activations at time step $t$. Thus ${y}_{t}^k$ is a score indicating the probability of $k^{th}$ label at time step $t$.

We utilize CTC transcription to determine the most likely sequence of class labels given $\mathbf{y}$.

\subsection{Training} \label{training}

Let the training dataset be denoted as $S={I_i,\mathbf{l}_i}$, where $I_i$ represents a word or line image and $\mathbf{l}_i$ represents its corresponding ground truth labeling. The objective function for training the encoder-decoder neural network for CTC transcription is derived from Maximum Likelihood principles. The aim is to minimize this objective function to maximize the log-likelihoods of the ground truth labeling. Therefore, the objective function utilized is:
\begin{equation}
   \mathbb{O}= - \sum_{I_i, \mathbf{l}_i \in S}  \log p (\mathbf{l}_i | \mathbf{y}_i),
\end{equation}
where $\mathbf{y}_i$ is the decoder output for the i\textsuperscript{th} sample. The above objective function can be optimized using gradient descent and back-propagation. 

\subsection{Inference} \label{inference}

During inference, the CTC-based classifier aims to output the labeling $\mathbf{l}^*$ with the highest probability, as defined in Eq.~(\ref{eq:p_of_l_given_x}). 
\begin{equation} \label{eq:p_of_l_given_x}
    p( \mathbf{l}|\mathbf{x}) = \sum_{\pi \in { \boldsymbol{\mathcal{B}}^{-1}(\mathbf{l})}} p(\pi|\mathbf{x}).
\end{equation}

\section{Experimental Setup} \label{sec:experimental_setup}
 
\subsection{Implementation Details} \label{sec:implementation_details}

In all experiments, cropped word or line images are resized to a height of 32 pixels and converted to grayscale, maintaining the original aspect ratio. To establish a validation split, we randomly select 5\% of pages from each book in the train split for all languages. It ensures that the validation split reflects the pages in the train split while the test split comprises pages from different sets of books. In Win\_RNN, the sliding window width $W_W$ is set to $20$, and the step size $W_S$ is set to $5$. For Col\_RNN, Win\_RNN, and CRNN, we utilize a bi-directional LSTM with 256 hidden units per direction across two layers, resulting in an output size of $2 \times 256$ at each time step. The CNN architecture in CNN\_only and CRNN follows the original CRNN paper~\cite{crnn_arxiv}. Our models are implemented using PyTorch~\cite{PYTORCH_NEURIPS}. We utilize an existing CRNN implementation~\cite{holmeyoung_crnn_github} for our experiments, conducting training on a single Nvidia GeForce 1080 Ti GPU. Training is set for 30 epochs. Word recognition models have a batch size of 64, while line recognition models use a batch size of 16. RMSProp~\cite{rmsprop} is employed as the optimizer. Col\_RNN and Win\_RNN are assigned a learning rate of $10e-03$, while CNN\_only and CRNN variants converge faster with a lower learning rate of $10e-04$.

\subsection{Evaluation}\label{sec:evaluation}

We need to assess text recognition in three scenarios: (i) word OCR: recognizing cropped word images, (ii) line OCR: recognizing cropped line images, and (iii) page OCR: end-to-end text recognition from document images. Our main evaluation metric in all cases is Character Accuracy (CA), determined by the Levenshtein distance between predicted and ground truth strings. For a formal definition of CA, let us denote the predicted text for a word/line/page as $l_i$ and the corresponding ground truth as $g_i$. If there are $N$ such samples, CA is defined as 
\begin{equation} \label{eq:charAcc_equation}
    CA= \frac{\sum_ilen(g_i) - \sum_i LD (l_i,g_i)}{\sum_ilen(g_i)} \times 100,
\end{equation}
where $len$ is a function that returns the length of the given string, and $LD$ is a function that computes the Levenshtein distance between the given pair of strings. Note that Character Error Rate (CER), another commonly used metric for OCR evaluation, is essentially $100 - CA$. We also include Sequence Accuracy (SA) alongside CA for word OCR and line OCR. SA represents the percentage of samples where the prediction is entirely correct (i.e., $LD(l_i,g_i)=0$). In the context of word recognition models, SA is equivalent to 'word accuracy' and is commonly used in scene text recognition literature. 

\begin{table}[!ht]
\center
\addtolength{\tabcolsep}{6.0pt}
\begin{tabular}{|l||r|r||r|r||r|r||r|r|} \hline
\textbf{Language} &\multicolumn{8}{c|}{\textbf{Word Recognition}}   \\\cline{2-9}
 &\multicolumn{2}{c||}{Col\_RNN} &\multicolumn{2}{c||}{Win\_RNN} &\multicolumn{2}{c||}{CNN\_only} &\multicolumn{2}{c|}{CRNN} \\\cline{2-9}
 &CA &SA &CA &SA &CA &SA &CA &SA \\\hline\hline
Assamese  &98.6  &95.4  &97.6  &92.9  &98.3  &96.0  &\textbf{99.0}  &\textbf{96.5} \\
Bengali  &99.1  &97.0  &98.3  &94.5  &99.2  &97.3  &\textbf{99.4}  &\textbf{97.9} \\
Guajrati  &96.2  &92.4  &95.1  &89.5  &96.2  &90.9  &\textbf{96.5}  &\textbf{93.9} \\
Hindi  &97.6  &95.1  &96.3  &92.3  &97.4  &94.2  &\textbf{98.2}  &\textbf{96.3} \\
Kannada  &97.4  &88.9  &96.4  &84.7  &96.7  &85.8  &\textbf{97.7}  &\textbf{90.7} \\
Malayalam  &99.5  &96.6  &99.3  &95.6  &98.0  &83.7  &\textbf{99.7}  &\textbf{97.7} \\
Manipuri  &98.6  &95.4  &97.8  &92.8  &98.2  &93.1  &\textbf{99.0}  &\textbf{96.9} \\
Marathi  &99.0  &96.2  &98.5  &94.2  &98.9  &95.0  &\textbf{99.2}  &\textbf{96.9} \\
Odia  &96.8  &93.5  &95.7  &90.8  &96.9  &93.7  &\textbf{97.2}  &\textbf{94.8}  \\
Punjabi  &99.1  &97.7  &98.4  &96.4  &99.2  &97.8  &\textbf{99.5}  &\textbf{98.7} \\
Tamil  &97.9  &91.0  &97.4  &88.4  &97.3  &87.2  &\textbf{98.0}  &\textbf{91.8} \\
Telugu  &96.3  &91.4  &95.3  &86.8  &96.4  &92.0  &\textbf{96.8}  &\textbf{93.6} \\
Urdu  &-  &-  &-  &-  &-  &-  &-  &-  \\\hline
\end{tabular}
\center
\caption{Results for recognition-only tasks are presented for each language individually on validation set of \textit{Mozhi} dataset. Each model configuration (Col\_RNN, Win\_RNN, CNN\_only, and CRNN) is trained separately for each language. Character Accuracy (CA) and Sequence Accuracy (SA) are reported for word recognition. The highest CA and SA values among the four encoder configurations are highlighted in bold.}\label{tab:internal_recogntion_only_val}
\end{table}

We employ a standard OCR evaluation toolkit for page OCR, where the input is a document image. Specifically, we utilize a modern adaptation~\cite{isri_modern} of the original ISRI Analytic Tools for OCR Evaluation~\cite{isri_1996}.
Using this toolkit, we compute Character Accuracy (CA) and Word Accuracy (WA). CA is calculated following the method described in Eq.~(\ref{eq:charAcc_equation}). Word accuracy is determined by aligning the sequences of words in the prediction $l_i$ with those in the ground truth $g_i$ and identifying the Longest Common Sub-sequence (LCS) between them. For a set of pages,
\begin{equation}
    WA =  \frac{\sum_i{len (LCS(l_i, g_i))}}{\sum_ilen(g_i)} \times 100
\end{equation}
where $len$ returns the number of words in a given sequence of words.

\section{Experiments and Results}
\label{sec:experiments}

\subsection{Comparing Different Encoder Configurations}

We assess the performance of four encoder configurations on the validation set of \textit{Mozhi} dataset for word recognition. Results are presented in Table~\ref{tab:internal_recogntion_only_val}. Each CA and SA pair in the table corresponds to a CTC-based network trained separately for a specific combination of language, recognition unit (word), and encoder configuration (Col\_RNN, Win\_RNN, CNN\_only, and CRNN). Across all cases except for Urdu word recognition, CRNN emerges as the top performer among the four configurations. The superior performance of CRNN over the CNN configuration highlights the necessity of capturing long-term dependencies in word or line images. Unlike fully connected networks, CNN layers have limited receptive fields, necessitating numerous layers to cover the entire input. Our seven-layer CNN lacks the depth to model extensive horizontal dependencies adequately. This deficiency is mitigated by employing a sequence encoder (bi-directional LSTM) that proficiently captures long-term dependencies in both directions. 

\begin{table}[!ht]
\center
\addtolength{\tabcolsep}{8.0pt}
\begin{tabular}{|l||rr||rr|}\hline
\textbf{Language}&\multicolumn{4}{c|}{\textbf{Test}} \\\cline{2-5}
 &\multicolumn{2}{c||}{Word}  &\multicolumn{2}{c|}{Line} \\\cline{2-5}
 &CA  &SA  &CA  &SA \\\hline\hline
Assamese  &98.9  &96.2  &99.2  &76.8 \\
Bengali   &99.0  &96.9  &98.1  &68.4 \\
Gujarati  &98.0  &94.9  &97.4  &63.1 \\
Hindi     &98.1  &95.5  &98.8  &63.5 \\
Kannada   &97.1  &88.7  &97.5  &53.9 \\
Malayalam &99.5  &97.3  &99.5  &87.3 \\
Manipuri  &98.4  &95.9  &99.2  &79.4 \\
Marathi   &99.3  &97.0  &99.3  &73.8 \\
Oriya     &97.5  &94.3  &98.8  &73.1 \\
Punjabi   &99.2  &98.2  &99.3  &79.7 \\
Tamil     &98.0  &91.6  &98.3  &68.1 \\
Telugu    &99.1  &95.4  &98.9  &71.7 \\
Urdu      &-  &-  &93.8  &24.2 \\\hline
\end{tabular}
\center
\caption{CRNN evaluation on test set of \textit{Mozhi} dataset. For each language, we train both word and line level CRNN models on the respective train split of the \textit{Mozhi} dataset.}\label{tab:public_data_results}
\end{table}
\begin{table}[!ht]
\addtolength{\tabcolsep}{6.0pt}
\center
\begin{tabular}{|l||rr|rr||rr|rr|}\hline
\textbf{Language}  &\multicolumn{4}{c||}{End-to-End OCR}  &\multicolumn{4}{c|}{GT Detection+CRNN }  \\\cline{2-9}
  &\multicolumn{2}{c|}{Tesseract} &\multicolumn{2}{c||}{Google} &\multicolumn{2}{c|}{GT word} &\multicolumn{2}{c|}{GT line} \\\cline{2-9}
&CA  &SA  &CA  &SA  &CA  &SA  &CA  &SA   \\\hline
Assamese  &90.0  &86.0  &92.7  &91.2 &99.3  &97.0  &99.4  &97.2 \\
Bengali   &91.3  &84.0 &96.2   &93.5 &99.1  &97.3  &99.0  &96.8 \\
Gujarati  &96.9  &92.4  &95.2 &93.0   &98.0  &93.7  &97.7  &91.9 \\
Hindi  &95.0  &93.3 &97.3 &95.2    &98.1  &96.0  &98.0  &95.6 \\
Kannada  &85.7  &84.6 &94.9  &85.1 &95.6  &89.2  &95.9  &86.4 \\
Malayalam &88.0  &74.8 &96.2  &89.7 &99.4  &98.0  &99.3  &97.9 \\
Manipuri  &85.7  &77.4 &90.9  &84.6 &98.4  &94.7  &98.7  &94.9 \\
Marathi  &97.9  &97.4  &98.4 &98.3    &99.6  &98.2  &99.5  &98.0 \\
Oriya  &94.0  &83.6  &92.6  &90.0  &98.6  &95.4  &98.0  &94.5 \\
Punjabi  &93.2  &89.8 &96.7 &92.7  &99.2  &98.3  &99.3  &97.9 \\
Tamil  &79.3  &42.4  &93.1 &92.5  &96.1  &85.6  &96.5  &85.4 \\
Telugu  &93.7  &79.3  &94.2  &89.2  &99.1  &95.1  &98.9  &94.0  \\
Urdu  &68.3  &26.2  &92.7  &85.7  &-  &-  &94.7  &81.5 \\\hline
\end{tabular}
\center
\caption{Performance of our page OCR pipelines compared to other public OCR tools. In this setting, we evaluate text recognition in an end-to-end manner on the test split of our dataset. Since the focus of this work is on text recognition, for end-to-end settings, for text detection, gold standard word/line bounding boxes are used. Under `End-to-End OCR' we show results of \textit{Tesseract}~\cite{tesseract_github} and \textit{Google Cloud Vision OCR}~\cite{google_cloud_vision}. Given a document image, these tools output a transcription of the page along with the bounding boxes of the lines and words detected. Under `GT Detection+CRNN', we show results of an end-to-end pipeline where gold standard word and line detection are used. For instance, 'GT Word' means we used ground truth (GT) word bounding boxes and the CRNN model trained for recognizing words, for that particular language. Bold value indicates the best result.}\label{table:page_level_end_to_end}
\end{table}

\subsection{Evaluating CRNN on Test Set of \textit{Mozhi}}

Table~\ref{tab:internal_recogntion_only_val} highlights that among four different models --- Col\_RNN, Win\_RNN, CNN\_only, and RCNN, RCNN obtained the best results for all languages on validation set of \textit{Mozhi} dataset with respect to CA and SA metrices for word recognition task. Since RCNN, highest performing model for validation set, we evaluated these models on test set of the same dataset. Table~\ref{tab:public_data_results} presents obtained results for word and line recognition on test set. 

\subsection{Page Level OCR Evaluation}

In page level OCR, the goal is to transcribe the text within a document image by segmenting it into lines or words and then recognizing the text at the word or line level. Our focus lies solely on text recognition, excluding layout analysis and reading order identification. To construct an end-to-end page OCR pipeline, we combine existing text detection methods with our CRNN models for recognition. Transcriptions from individual segments are arranged in the detected reading order. We evaluate the end-to-end pipeline by using gold standard detection to establish an upper bound on our CRNN model's performance. Additionally, we compare our OCR results with two public OCR tools: \textit{Tesseract} and \textit{Google Cloud Vision OCR}. Results from all end-to-end evaluations are summarized in Table~\ref{table:page_level_end_to_end}.

\begin{figure}[!ht]
\centerline{
\includegraphics[height=0.45\textwidth,width=0.45\textwidth]{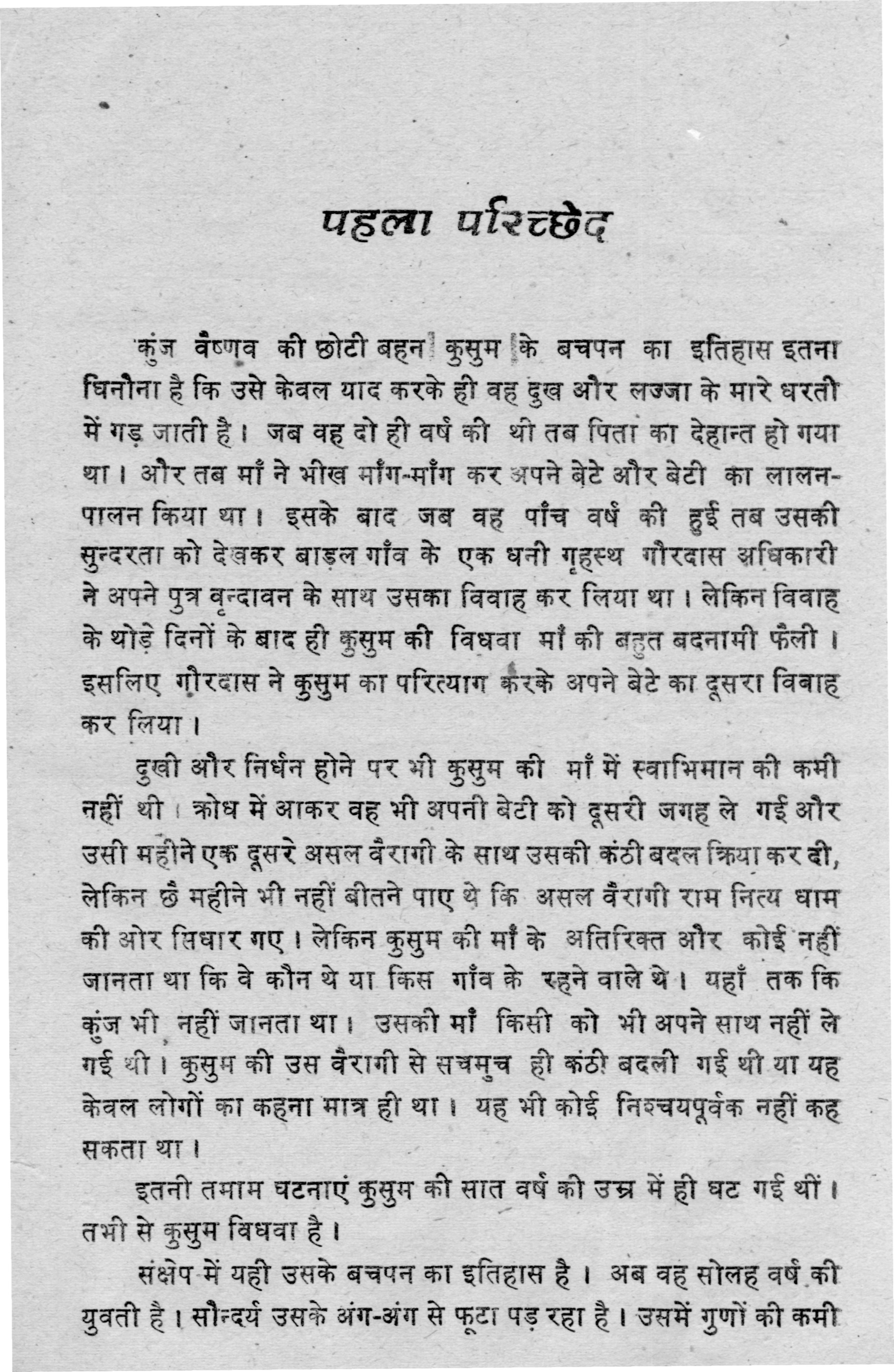}
\hspace{0.01\textwidth}
\includegraphics[height=0.45\textwidth,width=0.45\textwidth]{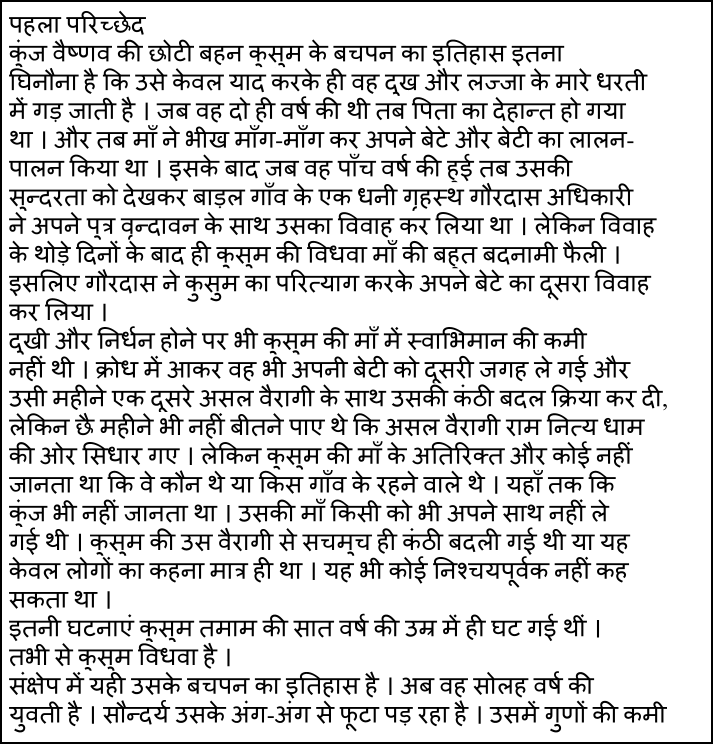}}
\centerline{(a)\hspace{0.5\textwidth}(b) }
\vspace{0.01\textwidth}
\centerline{
\includegraphics[height=0.45\textwidth,width=0.33\textwidth]{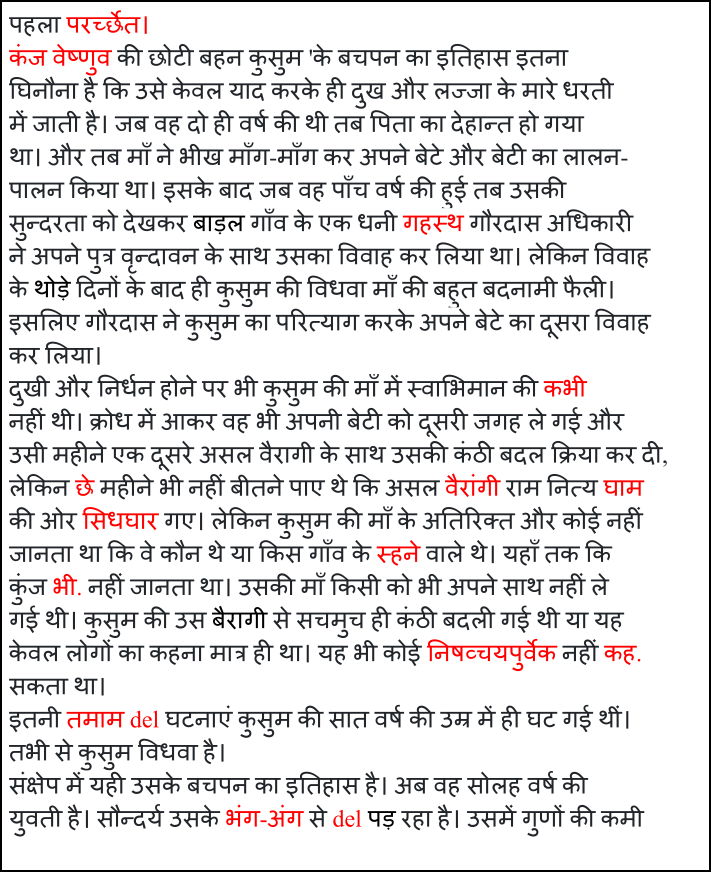}
\hspace{0.0001\textwidth}
\includegraphics[height=0.45\textwidth,width=0.33\textwidth]{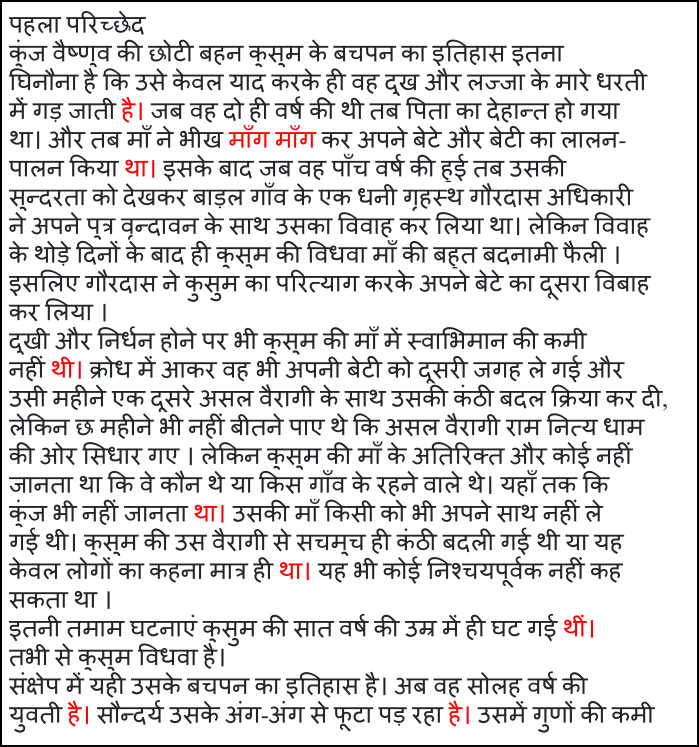}
\hspace{0.0001\textwidth}
\includegraphics[height=0.45\textwidth,width=0.33\textwidth]{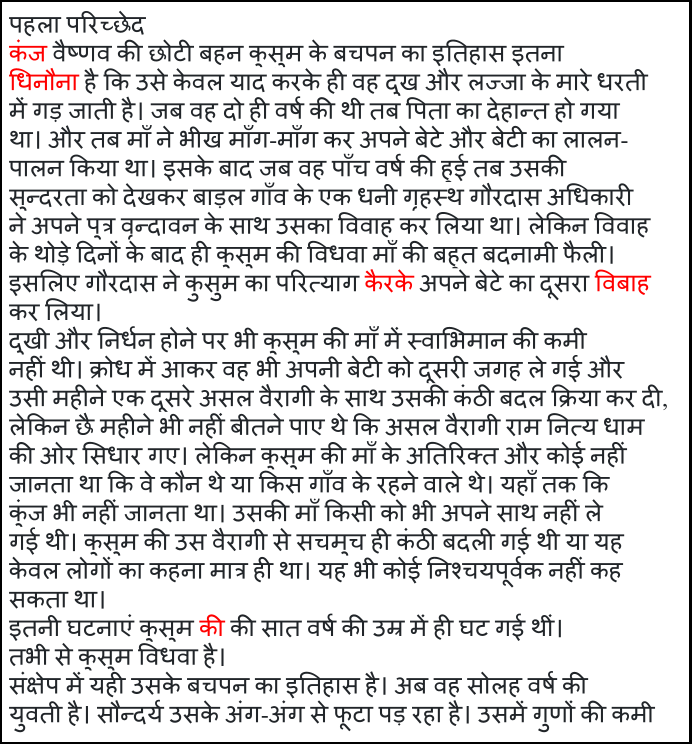}}
\centerline{(c)\hspace{0.3\textwidth}(d)\hspace{0.3\textwidth}(e)}
\caption{Displays qualitative results at the page level using \textit{Tesseract}, \textit{Google OCR}, and our method on a Hindi document image. For optimal viewing, zoom in. (a) original document image, (b) ground truth textual transcription, (c) predicted text by \textit{Tesseract}, (d) predicted text by \textit{Google OCR}, and (e) predicted text by our approach.}
\label{fig:hindi_page}
\end{figure}

In Fig.~\ref{fig:hindi_page}, visual results at the page level using \textit{Tesseract}, \textit{Google OCR}, and our approach are depicted. Panel (a) presents the original document image, while panels (b) to (e) display the ground truth and the predicted text by \textit{Tesseract}, \textit{Google OCR}, and our approach, respectively. Wrongly recognized texts are highlighted in red. This figure emphasizes that our approach outperforms existing OCR tools in producing accurate text outputs.

\subsection{Use Cases}

\begin{figure}
\centerline{
\includegraphics[height=0.3\textwidth,width=0.5\textwidth]{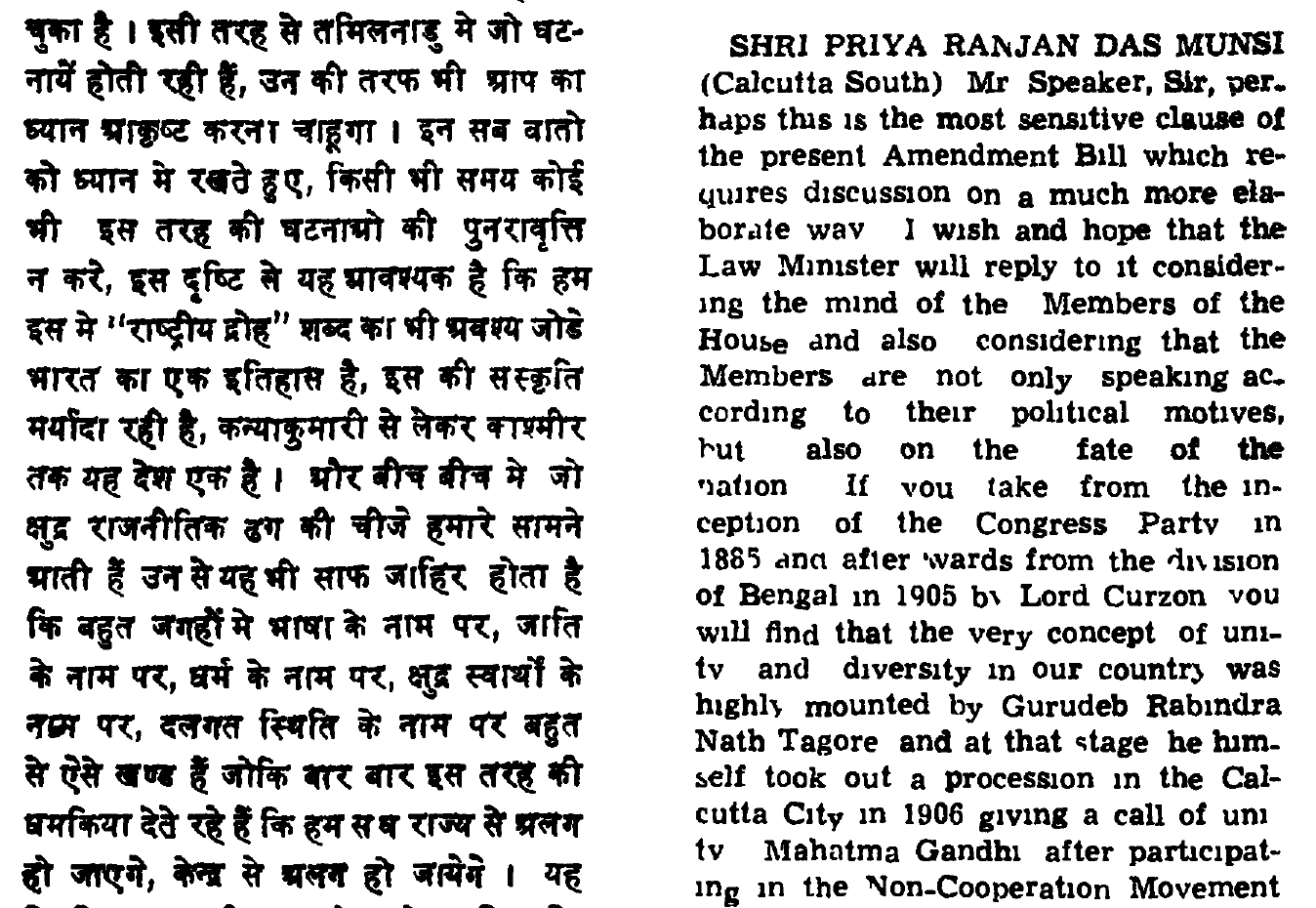}
\hspace{0.001\textwidth}
\includegraphics[height=0.3\textwidth,width=0.5\textwidth]{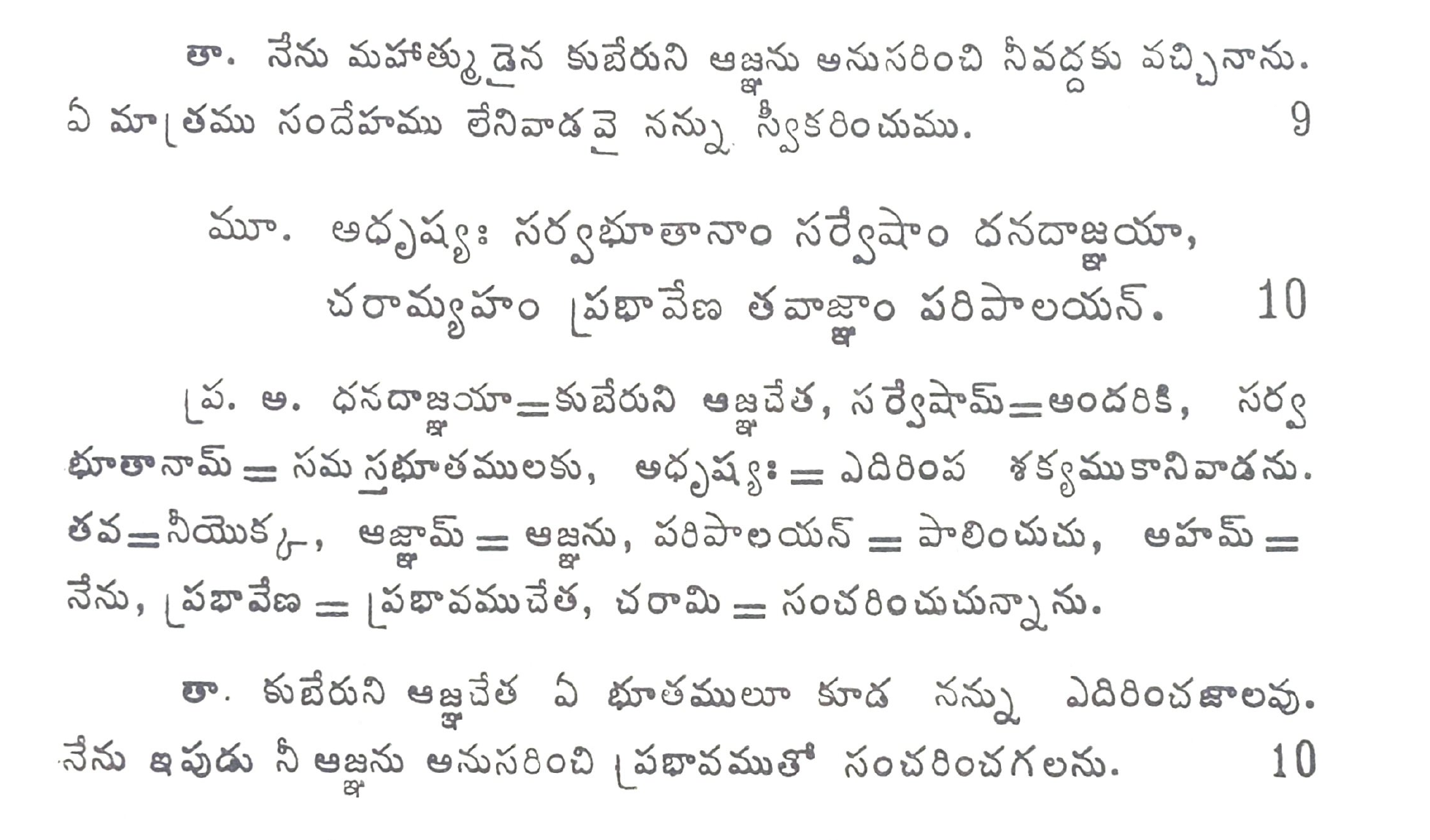}}
\centerline{(a) \hspace{0.4\textwidth} (b)}
\vspace{0.05\textwidth}
\centerline{
\includegraphics[height=0.35\textwidth,width=0.5\textwidth]{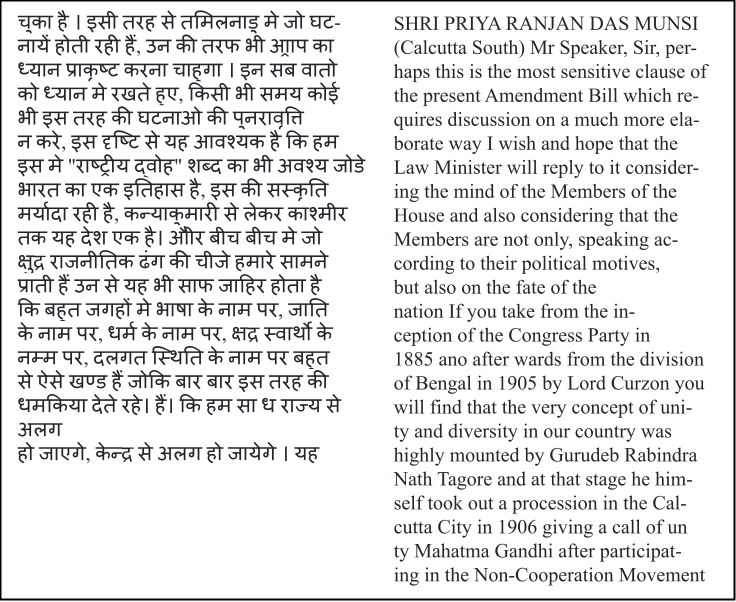}
\hspace{0.001\textwidth}
\includegraphics[height=0.35\textwidth,width=0.5\textwidth]{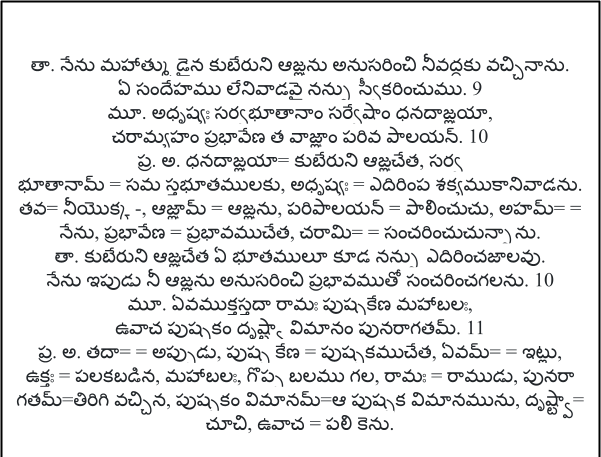}}
\centerline{(c)\hspace{0.4\textwidth} (d)}
\caption{Illustrates use cases for the digitization of \textit{Loksabha records} and \textit{Telugu Upanishad} pages. (a) and (b) display cropped regions from the original images of \textit{Loksabha} and \textit{Upanishad} documents, respectively. Panels (c) and (d) present the corresponding text outputs generated using our OCR APIs.}\label{fig:usecase}
\end{figure}

We leverage our OCR APIs for various significant applications. Notable examples include the pages of the \textit{Punjab Vidhan Sabha}, \textit{Loksabha records}, and \textit{Telugu Upanishads}. These digitization efforts enable easier access, preservation, and analysis of these valuable texts. The output and effectiveness of our OCR technology in these diverse use cases are illustrated in Fig.~\ref{fig:usecase}. These applications showcase the versatility and reliability of our OCR APIs in handling different scripts and document types, ensuring high accuracy and efficiency.

\subsection{Discussion}

Our method performs better in page level recognition than \textit{Tesseract} across all 13 languages, as evidenced by the results in Table~\ref{table:page_level_end_to_end}. Specifically, our approach surpasses \textit{Google} for eight languages, as indicated in the same table when considering ground truth bounding boxes. However, our dataset predominantly comprises pages from books, resulting in limited font, style, layout, and distortion diversity. Nevertheless, this dataset can serve as valuable pre-training data. Moving forward, we aim to enrich the dataset by gathering diverse documents with varying layouts, content, fonts, styles, and distortions, enhancing its comprehensiveness and utility for developing robust recognition models.

\section{Conclusions}

We empirically study different CTC-based word and line recognition models in 13 Indian languages. Our study concludes that CRNN, which uses a CNN for feature representation and a dedicated RNN-based sequential encoder, works best. Using existing text detection tools and our recognition models, we build page level OCR pipeline and show that our approach works better than two popular OCR tools for most of the languages. We also introduce a new public \textit{Mozhi} dataset for cropped word/line recognition in 13 Indian languages with more than 1.2 million annotated words. Additionally, we provide APIs for our page level OCR models and web-based applications that integrate these APIs to digitize Indic printed documents. We believe our study, the \textit{Mozhi} dataset, and available APIs will encourage research on OCR of Indian languages.

\section*{Acknowledgment}
This work is supported by MeitY, Government of India, through the NLTM-Bhashini project.

\bibliographystyle{splncs04}
\bibliography{mybibliography}

\end{document}